\documentclass{article}

\usepackage{nips_2018}

\usepackage[utf8]{inputenc} 
\usepackage[T1]{fontenc}    
\usepackage{hyperref}       
\usepackage{url}            
\usepackage{booktabs}       
\usepackage{amsfonts}       
\usepackage{amsmath}
\usepackage{nicefrac}       
\usepackage{microtype}      
\usepackage{graphicx}
\title{Visualizing Topographic Independent Component Analysis with Movies}

\begin{document}

\maketitle

\begin{abstract}
Independent component analysis (ICA) has often been used as a tool to model natural image statistics by separating multivariate signals in the image into components that are assumed to be independent.  However, these estimated components oftentimes have higher order dependencies, such as co-activation of components, that are not accounted for in the model.  Topographic independent component analysis (TICA), a modification of ICA, takes into account higher order dependencies and orders components topographically as a function of dependence.  Here, we aim to visualize the time course of TICA basis activations to movie stimuli.  We find that the activity of TICA bases are often clustered and move continuously, potentially resembling activity of topographically organized cells in the visual cortex.

\end{abstract}

\section{Introduction}
\subsection{Motivation}
Computational models of the early visual system have greatly informed our understanding of visual cortex organization. These models have linked known properties of the early visual system (e.g. motion selectivity, line orientation, contrast sensitivity) to the statistical properties of natural images (Barlow 1972; Field, 1994). One such computational model, known as independent component analysis (ICA), has been a tour de force in modelling natural image statistics. Similar to sparse coding (Olshausen and Field, 1997), ICA decomposes a multivariate signal into independent components that are localized in space and frequency. Implicit to ICA is the assumption of independence between components. However, as this assumption is oftentimes violated, a modification of the basic ICA model has been developed, which serves to model the shared dependencies between components. The spatially ordered topography of TICA-derived basis vectors has heralded this method as a biologically plausible model of early visual system properties. 

\subsection{Definition of ICA}
Independent component analysis (ICA) (Jutten and Herault, 1991) is a generative model that decomposes a multivariate signal into independent signals.  ICA can be represented by the linear model \\[0.1in] \centerline{\boldmath$x = As$}\\[0.1in]
\noindent where \boldmath$x$ $=(x_1, x_2, ... x_n)^T$ represents the observed data. \boldmath$A$ denotes the mixing matrix or basis matrix function, and \boldmath $s = (s_1, s_2, ... s_n) ^T$ describes the independent components of the data.  $A$ is unknown, and $s$ is a latent variable, which means that it cannot be directly observed.  Thus, both $A$ and $s$ need to be estimated from the data $x$. In order to do this, several assumptions need to be made about $A$ and $s$. First, the basic matrix function, $A$, is assumed to be square, full ranked, and typically sparse.  Second, the signal sources (independent components), $s$, are assumed to be independent of each other, or in other words, maximally non-gaussian. Using these assumptions, a learning rule can be derived that uses gradient descent on the negative log likelihood to estimate $A$. After convergence to a solution for $A$, $s$ can be computed by multiplying the observed data $x$ with the inverse of $A$.

\subsection{Definition of TICA}
Topographic Independent Component Analysis (TICA) (Hyvärinen, Hoyer, and Inki, 2001) is a modification of ICA that does not assume independence between the components $s$.  Instead, TICA considers the higher order dependencies between components by modelling the correlations of energies, or simultaneous activations of components. The advantage of this approach is that components can be organized topographically such that basis vectors that are spatially contiguous share higher dependencies while bases that are further apart share fewer dependencies (i.e. they are more independent). In ICA, the ordering of the bases is random; thus there is no topographic representation of components. TICA therefore provides a more intuitive way to visualize higher order dependencies of natural image statistics. 
\\[0.1in]
TICA assumes that variance controls the activation of components.  Thus, we can rewrite a component's variance as
\\[0.1in]
\centerline{$\sigma_i = \phi(\displaystyle\sum_{j=1}^{n} h(i, j)u_j)$}
\\[0.08in]
\centerline{where 
$h(i, j) = \begin{cases} 
1, if |i - j| \leq m \\
0, otherwise\end{cases} $}
\\[0.1in]

Here, we define $h(i, j)$ as a neighborhood function that describes the distance in topography between the i-th and j-th components as a function of neighborhood width, $m$. If components $i$ and $j$ are within the same topographic neighborhood, then $h(i, j) = 1$. If the distance is otherwise larger than the width of the neighborhood,  $h(i, j) = 0$. The variance can be used to obtain $s$ using the equations 
\\[0.1in]
\centerline{$s_i = \sigma_i z_i$} \\[0.08in]
\centerline{$s_i =  \phi((\displaystyle\sum_{j=1}^{n}  h(i, j)u_j)z_i$}
\\[0.1in]
where $z_i$ is a random variable with the same distribution as $s_i$.  These equations can then be used to derive the learning rule estimating $A$, to determine topographic dependencies. The derivation of TICA's learning rule has been described in detail in Hyvärinen, Holyer, and Inki's 2001 paper (equations 21-27).
\subsection{Our Contribution}
Bases that have been trained on natural images hint at the emergent complex topography of neurons in the early visual system. Interestingly, while static images of natural scenes have dominated the TICA literature, movie clips have not yet been explored as a useful alternative. From a computational standpoint, movies are also likely to be useful as they allow for real-time visualization of the time course co-activations of neighboring basis vectors as objects move across the screen. In light of this gap, we have implemented the first systematic investigation of TICA on simple and complex movies and we demonstrate the potential for this approach to better inform our understanding of the properties of the early visual system. 

\section{Methods}

\subsection{Obtaining training data from natural images}

The training data were obtained in line with the methods reported by Hyv{\"a}rinen, Hoyer, and Inki (2001). Gray-scaled natural scenes depicting animals, trees, natural scenes were obtained from PhotoCD \footnote{http://www.cis.hut.fi/projects/ica/data/images/} (size: 256 x 512 or 512 x 256 pixels). Each image was normalized with zero mean and unit variance. The training data were derived from 100x100 pixel image patches at random locations from the normalized images. Each image was converted into a vector of length 10,000 and the mean gray-scale value was subtracted from each image vector. Principal component analysis (PCA) was then used to reduce the dimension of the data vector while simultaneously pre-whitening. Specifically, we retained the first 200 principal components (PCs) explaining the largest amount of variance and then normalized the variance of these principal components:
\\[0.1in]
\centerline{$ z = Vx$}
\\[0.1in]
The whitening matrix $ V $ is computed as $ V = D^{-1/2}E^T$, where $ D $ and $ E $ are defined by the eigenvalue decomposition of $ E\{ xx^T \} = EDE^T $. The same whitening matrix obtained here is used to whiten the test data, and the inverse of this whitening matrix is used to recover the basis vectors and the reconstructed images. The number of PCs determine the number of basis vectors (200) in the learned topographical map. 

\subsection{Estimating the TICA model} 
This work is based on a publically available implementation of TICA by Hoyer and Hyvärinen\footnote{https://research.ics.aalto.fi/ica/imageica/}. A 2-D torus lattice with a neighborhood size of 3 x 3 square of 9 units defined the topographical map (Kohonen, 1995) of width 20 and height 10 (to encompass 200 PCs). To update the weight vectors, we used the same gradient learning rule and adaptive step size (see equations 27 and 28) as described by Hyv{\"a}rinen, Hoyer, and Inki (2001). The starting step size was 0.1 and the constant parameter $\epsilon$ was 0.005. After learning the $ w_i $, the original mixing matrix $ A $ was computed by inverting the whitening process as
\\[0.1in]
\centerline{$ A = (WV)^{-1} = V^{-1}W^T $}
\\[0.1in]
where the inverse of A contains bases in the original space but not the whitened space.

\subsection{Obtaining test data from movies} 

For the test data, movie clips were taken from Hollywood movies (e.g. "Birdman"\footnote{https://www.youtube.com/watch?v=cX3VbKEooWQ}) filmed with the long-take technique. This approach ensured that there were no abrupt scene transitions by maximizing continuity across frames. The cinematic black bars on the top and the bottom of the movie frames were removed and the frame images were gray-scaled and re-sized to a length of 500 pixels. The testing data were obtained by taking 100x100 pixel image patches at the same location across frames in the same movie clip. These images were then processed using the same pipeline adopted for the training data (see section 2.1). The dimensions of the movie test data were reduced by multiplying the frame data with the whitening matrix $ V $ obtained from the training data.

\subsection{Visualizing basis activation and reconstructed images} 
In order to place the activation values in whitened space, the inverse of $ A $ was multiplied with the whitened movie test data $z$. Then, the inverse of the whitening matrix $V$ was multiplied with the whitened activation matrix to recover the activation in the original space.
\\[0.1in]
\centerline{$ s = V^{-1} A^{-1} z$}
\\[0.1in]
Basis energy was obtained by calculating the square of activation ($ s^2 $) and the reconstructed images $\hat{x}$ were derived from the basis vectors by multipling the mixing matrix $A$ with the activation $s$ as 
\\[0.1in]
\centerline{$ \hat{x} = As$}
In our visualization videos, we slowed down the speed of the movie clips by 3x for better visualization.

\section{Results}
\subsection{Validation of the activation with a single basis}
To confirm that we obtained the basis activations correctly, we used the image of a single basis from the matrix $ A $ as the test data and visualized its activation (see Figure 1A). A video of the visualization in real-time is publically available \footnote{https://www.youtube.com/watch?v=ORCWe2ZVVuc}. We found that each corresponding basis vector had a high activation while all other vectors displayed close to zero activation. This result confirmed that our visualization of basis activation was correct, as well as demonstrating that the basis vectors were uncorrelated with each other.

\begin{figure}
\begin{center}
\includegraphics[width=1\textwidth]{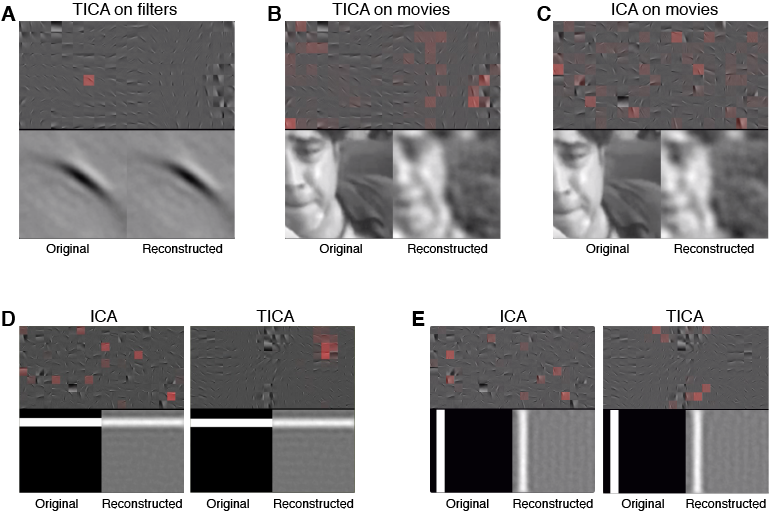} 
\caption{A) TICA basis energy for a single filter. B) TICA basis energy for a 100 x 100 image patch from the movie "Birdman". c) ICA basis energy for the same image patch as in B. D) Basis energy comparison between ICA and TICA on a simple horizontal bar stimulus. E) Comparison between ICA and TICA for a simple vertical bar stimulus.}
\label{fig:single-basis}
\end{center}
\end{figure}

\begin{figure}[h!]
\begin{center}
\includegraphics[width=0.65\textwidth]{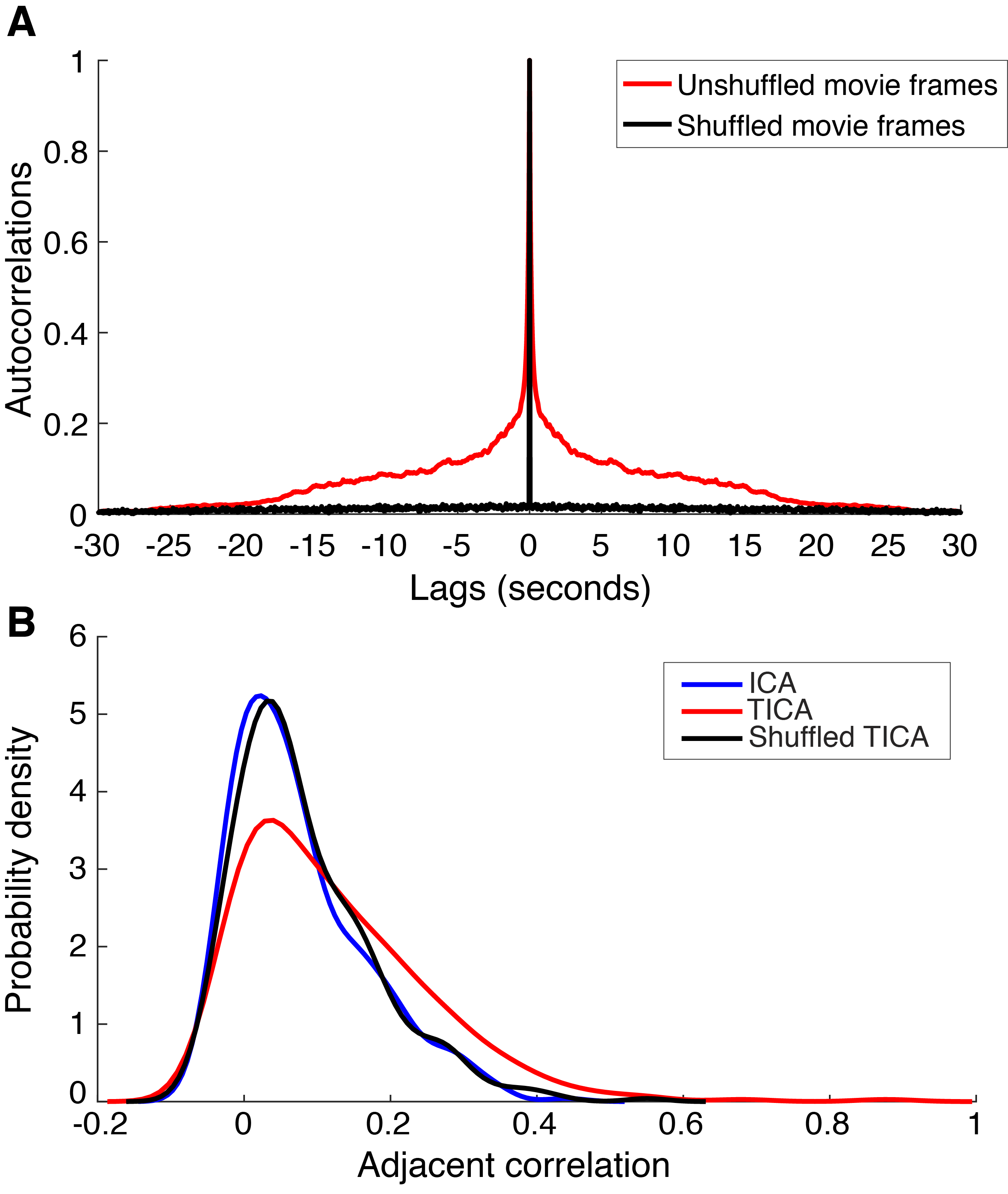}
\caption{A) Autocorrelation for basis energy. B) Adjacent correlation for basis energy.}
\label{fig:autocorr-energy}
\end{center}
\end{figure}

\subsection{Basis activation for a moving bar}
Moving bars are often used to quantify the response characteristics of neurons in the visual cortex (Hubel and Wiesel, 1962; Dumoulin and Wandell, 2008). Since the spatial arrangement of the bases in topographic ICA resembles that of V1 complex cells, simulating the activations of the bases in response to moving gratings may provide us with useful insights into the organization of upstream areas of the visual cortex (typically sensitive to motion).

We visualized the basis activation energies in response to a moving bar at varying orientations (Figure 1D and 1E). We found that with TICA, these moving bars activated spatially localized clusters of bases, which moved continuously as the bar moved\footnote{https://www.youtube.com/watch?v=V28P-gliDuM}. This finding is consistent with the idea that the topographic arrangement of the bases allows for more efficient wiring of motion-detecting neurons in higher layers of the visual cortex. For comparison, the same visualizations were performed on bases obtained by ICA\footnote{https://www.youtube.com/watch?v=kz76K32-INU}. For ICA, we did not observe such localization of activations or continuous changes over frames.

\subsection{Visualizing basis activation on movies}

In order to simulate basis activations of more naturalistic stimuli, we used natural movies. We cropped 100x100 pixel frames from the movie "Birdman" and slowed down the frames by 3x for the purpose of visualizing the activations more easily\footnote{https://www.youtube.com/watch?v=S0xFoolBKX4}. For frames which could be captured by a few bases, there was usually a clear spatial progression of activity from one adjacent basis to the next. For frames which activated many bases simultaneously, it became harder to notice the progression of activation between adjacent cells. 

For comparison, we also visualized basis activations using ICA and shuffled TICA. The spatial clustering of activations, observed with TICA, was lost with ICA\footnote{https://www.youtube.com/watch?v=HN2f0Te7sqk}. When the TICA frames were shuffled, the temporal correlation and spatial clustering of activations was lost\footnote{https://www.youtube.com/watch?v=fd9k3bmZsMs}.

In the following sections, we describe two correlational analyses in an attempt to quantify the activation of the bases over movie frames.

\subsubsection{Autocorrelation of basis activation}
We investigated the temporal and spatial profiles of basis activations. In the temporal domain, we quantified how stable these activations were by computing the autocorrelation of the activation for each basis over the frames of the movie (Figure 2A in red). As a comparison, we shuffled the order of the movie frames and calculated the autocorrelation of the activations (Figure 2A in black). There was consistent non-zero autocorrelation in basis activations up to about 20 seconds. This demonstrates that the TICA bases were capturing temporal features of the movies. Surprisingly, the timescale of this autocorrelation is similar to the behavioral data showing that perception of oriented bars are biased to recently seen stimuli as far as ~15 seconds ago (Fischer and Whitney, 2011). The timescale that the autocorrelation was significant for may provide an indication of the timescales at which the basis activations propagate over bases.

\subsubsection{Adjacent correlation of basis energy}
To quantify the magnitude of dependency between adjacent units, we calculated the Pearson correlation of the energies, $s^2$, between adjacent units. We compared the distribution of these correlations for bases obtained by ICA, TICA and shuffled TICA bases. The mean adjacent correlation for TICA (R=0.123) was higher than for ICA (R=0.076). A random permutation test produced a p value of < 0.001 of finding such a difference in means by chance. There was no significant difference between the correlations for ICA (R=0.076) and shuffled TICA (R=0.083), with a permutation test producing a p value of 0.297. These findings are consistent with our qualitative observations of the basis activations in both the grating and movie clips: bases with high activations tended to be clustered together.

\section{Conclusion}
Here, we systematically explored the activation of TICA bases on movie clips. To do this, we derived basis vectors on natural images and then we tested these vectors on stimuli varying in complexity, from simple gratings to naturalistic scenes. We found that movies are a very useful way of visualising the simultaneous co-activations of topographically ordered basis vectors, in both the spatial and temporal domain. Further corroborating this result, we found that ICA did not yield co-activations of basis vectors in the spatial domain, reflecting the fact that the bases derived using this method do not contain any meaningful topographical structure. In sum, our results confirm the biological tractability of TICA for exploring early visual system properties. Future research would benefit by comparing these basis activations to populations of neuronal activations in the visual cortex. This effort would serve to further link together computational models to biological properties of the early visual system.

\section*{Acknowledgements}
We would like to thank Bruno Olshausen for his guidance and support throughout the project.

\section*{References}
Barlow, H. B. (1972). Single units and sensation: a neuron doctrine for perceptual psychology? Perception, 1(4), 371–394.
\\[0.12in]
Dumoulin, S. O., \& Wandell, B. A. (2008). Population receptive field estimates in human visual cortex. NeuroImage, 39(2), 647–660.
\\[0.12in]
Field, D. J. (1994). What Is the Goal of Sensory Coding? Neural Computation, 6(4), 559–601.
\\[0.12in]
Fischer, J., \& Whitney, D. (2014). Serial dependence in visual perception. Nature neuroscience, 17(5), 738.
\\[0.12in]
Hubel, D. H., \& Wiesel, T. N. (1962). Receptive fields, binocular interaction and functional architecture in the cat’s visual cortex. The Journal of Physiology, 160(1), 106–154.
\\[0.12in]
Hyvärinen, A., Hoyer, P. O., \& Inki, M. (2001). Topographic independent component analysis. Neural Computation, 13(7), 1527–1558.
\\[0.12in]
Jutten, C., \& Herault, J. (1991). Blind separation of sources, part I: An adaptive algorithm based on neuromimetic architecture. Signal Processing, 24(1), 1–10.
\\[0.12in]
Kohonen, T. (1995). Learning Vector Quantization. In T. Kohonen (Ed.), Self-Organizing Maps (pp. 175–189). Berlin, Heidelberg: Springer Berlin Heidelberg.
\\[0.12in]
Olshausen, B. A., \& Field, D. J. (1997). Sparse coding with an overcomplete basis set: a strategy employed by V1? Vision Research, 37(23), 3311–3325.
\\[0.12in]
Tootell, R. B., Switkes, E., Silverman, M. S., \& Hamilton, S. L. (1988). Functional anatomy of macaque striate cortex. II. Retinotopic organization. The Journal of Neuroscience: The Official Journal of the Society for Neuroscience, 8(5), 1531–1568.
\\[0.12in]
\end{document}